\documentclass[]{ceurart}

\begin{document}
\pdfoutput=1
\copyrightyear{2022}
\copyrightclause{Copyright for this paper by its authors.
  Use permitted under Creative Commons License Attribution 4.0 International (CC BY 4.0).}

\conference{LM-KBC'22: Knowledge Base Construction from Pre-trained Language Models,
  Challenge at ISWC 2022}

\title{Task-specific Pre-training and Prompt Decomposition for Knowledge Graph Population with Language Models}

\author[1]{Tianyi Li}[%
email=tianyi.li@ed.ac.uk]
\author[1]{Wenyu Huang}[%
email=s2088661@ed.ac.uk]
\author[2]{Nikos Papasarantopoulos}[%
email=nikos.papasarantopoulos@huawei.com]
\author[2]{Pavlos Vougiouklis}[%
email=pavlos.vougiouklis@huawei.com]
\author[1,2]{Jeff Z. Pan}[%
url=http://knowledge-representation.org/j.z.pan/]

\address[1]{ILCC, School of Informatics, University of Edinburgh, United Kingdom}
\address[2]{Huawei Edinburgh Research Centre, United Kingdom}

\begin{abstract}
  We present a system for knowledge graph population with Language Models, evaluated on the Knowledge Base Construction from Pre-trained Language Models (LM-KBC) challenge at ISWC 2022. Our system involves task-specific pre-training to improve LM representation of the masked object tokens, prompt decomposition for progressive generation of candidate objects, among other methods for higher-quality retrieval. Our system is \textbf{the winner of track 1 of the LM-KBC challenge}, based on BERT LM; it achieves 55.0\% F-1 score on the hidden test set of the challenge.\footnote{Our code and data are available at \url{https://github.com/Teddy-Li/LMKBC-Track1}.}
\end{abstract}

\maketitle

\section{Introduction}

Knowledge graph population is a task of predicting the objects from given subject-relation pairs. For example, for the subject-relation pair $<Andalusia, \\ \textit{StateSharesBorderState}>$, the task is to predict the appropriate objects such as \textit{Faro}, \textit{Beja}, \textit{Gibraltar}, etc . The task of knowledge graph population is highly related to the task of link prediction in the knowledge graph and Natural Language Processing (NLP) literature \cite{socher_reasoning_2013,bordes_translating_2013}; the key difference is that, in knowledge graph population the objects are generated not from a fixed pool of entity nodes, but from an open vocabulary of words.

Knowledge graph population plays an important role in knowledge graph construction: given a relation ontology and a set of seed subject entities, knowledge graph population can produce a graph of relation triples that is not restricted to any given entity sets.

In this work, we develop our system on the LM-KBC challenge at ISWC 2022, a challenge of knowledge graph population concerning 12 relations. In this challenge, each system is provided with subject-relation pairs, and asked for the appropriate objects of these subject-relation pairs. Each subject-relation pair can be associated with zero, one, or multiple true objects, echoing the complex situation in real-world scenarios.

Our system falls in the \textbf{track 1} of the challenge: it seeks to improve BERT \cite{devlin_bert_2019} language model's performance in knowledge graph population from the following three dimensions: 1) LM representation of masked object tokens; 2) candidate object generation; 3) candidate object selection (ranking). For improving LM representations, we apply task-specific pre-training, utilizing silver data retrieved from Wikidata \cite{noauthor_wikidata_nodate} to aid the training process; for candidate generation, we use prompt decomposition to convert complex knowledge graph population tasks into multiple simpler tasks; for candidate selection, we use adaptive thresholds, with explorations to methods for relaxing the single-true-object assumption behind Softmax normalization.

In comparison to the winning submission in track 2 \cite{alivanistos_prompting_2022}, we highlight the following contributions: 1) we show the effectiveness of task-specific pre-training, particularly when doing so separately for each individual relation; 2) we propose to decompose prompts to split the task into multiple steps, in order to achieve best results under the constraint of LM size and capability.

Below, we discuss the above three dimensions of improvements in details in Section \ref{Sec:pretraining}, \ref{Sec:candidate_generation}, \ref{Sec:candidate_selection}, then describe our main experiment results in \ref{Sec:experiments}.

Our method is based on BERT-large-cased LM\footnote{https://huggingface.co/bert-large-cased}, since as a general observation we have found cased BERT models to outperform uncased ones; we speculate that this can be attributed to an explicit distinction between named entities and general nouns by capitalization of the first characters. For all supervised experiments, we split the train set further into a train2 and a dev2 set with respective portions of 80\% and 20\%. We use the train2 set for training and the dev2 set for checkpointing; this way the dev set is kept as a hidden evaluation dataset, on which we report results through sections \ref{Sec:pretraining} to \ref{Sec:candidate_selection}.

\section{Task-Specific Pre-training for Better Representations}
\label{Sec:pretraining}

Language models like BERT \cite{devlin_bert_2019} have been trained on diverse texts in large scale; therefore, by ``reminding'' the models of what type of information they are supposed to recall from pre-training, performance is expected to be improved. Along this line, \cite{gururangan_dont_2020} have shown that adaptive pre-training is helpful for language models' performance on target domains. In our system, we explore the approach of training the BERT language model under MLM objective with the subject-relation-object triples.

The task-specific pre-training approach can be summarized as follows: given a subject-relation-object triple, we use the triple to instantiate the corresponding prompt template, to create a sentence. We then mask those tokens in the sentence that are relevant to the object entity, and train the BERT models with the masked sentence, where the objective is to recover these masked tokens.

One interesting dimension of freedom in this task is which tokens to mask. This is motivated by our end task, to predict the tokens in the place of objects. Therefore, the representation of those object tokens is what we are most keen on improving. We further hypothesize that improving the representation of tokens close to the object tokens\footnote{For instance, the tokens ``a'' and ``.'' in the sentence ``A cat sits on a mat .''} may also help with the prediction of object tokens. Thus, in summary, we mask the tokens corresponding to the object, as well as tokens beside the object tokens, up to a window of $c$ tokens on each side.

\begin{table}[h!]
    \centering
    \caption{Dev set results on all relations, with raw BERT-large-cased model or with intermediate pre-training with $c=2$. Highlighted blocks have the best performance of the four configurations.}
    \begin{tabular}{|c|c|c|c|c|c|c|}
        \hline
        \multirow{2}{*}{Relation} & \multicolumn{3}{|c|}{Raw BERT} &  \multicolumn{3}{|c|}{MLM $c=2$} \\\cline{2-7}
         & Precision & Recall & F-1 & Precision & Recall & F-1 \\\hline
        ChemicalCompoundElement   & 0.624 & 0.530 & 0.542 & 0.777 & 0.737 & 0.736 \\
        CompanyParentOrganization & 1.000 & 0.680 & 0.680 & 1.000 & 0.680 & 0.680 \\
        CountryBordersWithCountry & 0.624 & 0.528 & \textcolor{red}{0.533} & 0.322 & 0.418 & 0.327 \\
        CountryOfficialLanguage   & 0.936 & 0.749 & \textcolor{red}{0.797} & 0.894 & 0.739 & 0.775 \\
        PersonCauseOfDeath        & 1.000 & 0.500 & 0.500 & 1.000 & 0.500 & 0.500 \\
        PersonEmployer            & 0.022 & 0.035 & \textcolor{red}{0.025} & 0.034 & 0.025 & 0.018 \\
        PersonInstrument          & 0.560 & 0.504 & 0.509 & 0.417 & 0.804 & 0.396 \\
        PersonLanguage            & 0.740 & 0.671 & 0.670 & 0.810 & 0.710 & 0.716 \\
        PersonPlaceOfDeath        & 1.000 & 0.500 & 0.500 & 1.000 & 0.500 & 0.500 \\
        PersonProfession          & 0.101 & 0.239 & 0.127 & 0.623 & 0.552 & \textcolor{red}{0.546} \\
        RiverBasinsCountry        & 0.626 & 0.560 & \textcolor{red}{0.547} & 0.149 & 0.193 & 0.145 \\
        StateSharesBorderState    & 0.227 & 0.153 & 0.162 & 0.173 & 0.136 & 0.140 \\\hline
        \textbf{Average}             & 0.510 & 0.469 & 0.464 & 0.597 & 0.499 & 0.455 \\\hline
    \end{tabular}
    \label{tab:mlm_1}
\end{table}

\begin{table}[h!]
    \centering
    \caption{Dev set results for all relations, continuation of Table \ref{tab:mlm_1}, with intermediate pre-training with $c=1$ or $c=0$. Highlighted blocks have the best performance of the four configurations.}
    \begin{tabular}{|c|c|c|c|c|c|c|}
        \hline
        \multirow{2}{*}{Relation} & \multicolumn{3}{|c|}{MLM $c=1$} & \multicolumn{3}{|c|}{MLM $c=0$} \\\cline{2-7}
         & Precision & Recall & F-1 & Precision & Recall & F-1 \\\hline
        ChemicalCompoundElement   & 0.807 & 0.679 & 0.706 & 0.825 & 0.712 & \textcolor{red}{0.745} \\
        CompanyParentOrganization & 1.000 & 0.680 & 0.680 & 1.000 & 0.680 & 0.680 \\
        CountryBordersWithCountry & 0.292 & 0.506 & 0.312 & 0.414 & 0.473 & 0.388 \\
        CountryOfficialLanguage   & 0.883 & 0.765 & 0.792 & 0.915 & 0.739 & 0.783 \\
        PersonCauseOfDeath        & 1.000 & 0.500 & 0.500 & 1.000 & 0.500 & 0.500 \\
        PersonEmployer            & 0.010 & 0.010 & 0.010 & 0.012 & 0.015 & 0.012 \\
        PersonInstrument          & 0.543 & 0.694 & 0.415 & 0.820 & 0.603 & \textcolor{red}{0.580} \\
        PersonLanguage            & 0.810 & 0.752 & \textcolor{red}{0.742} & 0.790 & 0.673 & 0.683 \\
        PersonPlaceOfDeath        & 0.420 & 0.520 & 0.180 & 0.540 & 0.520 & 0.240 \\
        PersonProfession          & 0.655 & 0.532 & 0.542 & 0.627 & 0.542 & 0.538 \\
        RiverBasinsCountry        & 0.193 & 0.288 & 0.200 & 0.483 & 0.336 & 0.340 \\
        StateSharesBorderState    & 0.241 & 0.161 & \textcolor{red}{0.165} & 0.177 & 0.113 & 0.128 \\\hline
        \textbf{Average}             & 0.568 & 0.507 & 0.435 & 0.630 & 0.496 & \textcolor{red}{0.467} \\\hline
    \end{tabular}
    \label{tab:mlm_2}
\end{table}

\begin{table}[h!]
    \centering
    \caption{Sizes of additional training data retrieved from Wikidata.}
    \begin{tabular}{|c|c|}
        \hline
        Relation & Size of Additional Data \\\hline
        ChemicalCompoundElement   & 3431 \\
        CompanyParentOrganization & NA \\
        CountryBordersWithCountry & NA \\
        CountryOfficialLanguage   & NA \\
        PersonCauseOfDeath        & NA \\
        PersonEmployer            & 8768 \\
        PersonInstrument          & 9359 \\
        PersonLanguage            & 1706 \\
        PersonPlaceOfDeath        & NA \\
        PersonProfession          & NA \\
        RiverBasinsCountry        & 278 \\
        StateSharesBorderState    & 1595 \\\hline
    \end{tabular}
    \label{tab:mlm_sizes}
\end{table}

Another interesting dimension of freedom is what training data to use. As a baseline we have the training set from the LM-KBC challenge to use for generating sentences; however, the scale of the training set (i.e. 100 subjects per relation) is very small even for a fine-tuning dataset. To mitigate this data sparsity issue, we further refer to Wikidata \cite{noauthor_wikidata_nodate} for more data entries, collecting the set of subject-object pairs recorded in Wikidata as satisfying each relation. Notably, to maintain the integrity of the evaluation, we exclude any overlaps between the subjects in any subset of the challenge from the subjects in the retrieved entity-pairs\footnote{We exclude entries by subject because: 1) it is more secure to exclude subject mentions with arbitrary objects than to exclude subject-object pairs; 2) for the challenge test set only the subjects are available, so by excluding overlaps by subjects, we ensure our models are \textbf{not} peeking the test set in any way.}.

\begin{table}[h!]
    \centering
     \caption{Dev set results for all relations, with intermediated pre-training with silver data concatenated to gold train2 set. ``joined'' refers to training one single checkpoint with training data for all relations; ``separate'' means each relation has its own checkpoint, trained with data only from this relation. For ``separate'' checkpoints, ``CompanyParentOrganization'', ``PersonCauseOfDeath'' and ``PersonPlaceOfDeath'' are not experimented with because they cannot outperform null strategy; ``PersonProfession'' is not experimented with because its SPARQL query on Wikidata exceeded the 1 minute time limit; ``CountryOfficialLanguage'' is not experimented with because there are just around 200 countries / regions in total.}
    \begin{tabular}{|c|c|c|c|c|c|c|}
        \hline
        \multirow{2}{*}{Relation} & \multicolumn{3}{|c|}{MLM silver joined $c=0$} & \multicolumn{3}{|c|}{MLM silver separate $c=0$ } \\\cline{2-7}
         & Precision & Recall & F-1 & Precision & Recall & F-1 \\\hline
        ChemicalCompoundElement   & 0.799 & 0.855 & 0.805 & 0.897 & 0.833 & \textcolor{red}{0.839} \\
        CompanyParentOrganization & 1.000 & 0.680 & 0.680 & NA & NA & NA \\
        CountryBordersWithCountry & 0.611 & 0.516 & 0.508 & NA & NA & NA \\
        CountryOfficialLanguage   & 0.837 & 0.762 & 0.753 & NA & NA & NA \\
        PersonCauseOfDeath        & 1.000 & 0.500 & 0.500 & NA & NA & NA \\
        PersonEmployer            & 0.032 & 0.065 & 0.038 & 0.322 & 0.035 & 0.025 \\
        PersonInstrument          & 0.380 & 0.444 & 0.232 & 0.820 & 0.383 & \textcolor{red}{0.325} \\
        PersonLanguage            & 0.143 & 0.177 & \textcolor{red}{0.133} & 0.133 & 0.146 & 0.119 \\
        PersonPlaceOfDeath        & 1.000 & 0.500 & 0.500 & NA & NA & NA \\
        PersonProfession          & 0.6146 & 0.561 & 0.553 & NA & NA & NA \\
        RiverBasinsCountry        & 0.657 & 0.557 & 0.571 & 0.647 & 0.628 & \textcolor{red}{0.584} \\
        StateSharesBorderState    & 0.247 & 0.191 & 0.187 & 0.280 & 0.184 & \textcolor{red}{0.198} \\\hline
        \textbf{Average}             & 0.613 & 0.484 & 0.455 & NA & NA & NA \\\hline
    \end{tabular}
    \label{tab:mlm_3}
\end{table}

\begin{table}[h!]
    \centering
    \caption{Dev set results for all relations, from a joint effort of the family of LM checkpoints.}
    \begin{tabular}{|c|c|c|c|c|}
        \hline
        \multirow{2}{*}{Relation} & \multirow{2}{*}{Checkpoint} &  \multicolumn{3}{|c|}{Result} \\\cline{3-5}
         & & Precision & Recall & F-1 \\\hline
        ChemicalCompoundElement   & MLM silver separate & 0.897 & 0.833 & 0.839 \\
        CompanyParentOrganization & raw BERT-large-cased & 1.000 & 0.680 & 0.680 \\
        CountryBordersWithCountry & raw BERT-large-cased & 0.608 & 0.559 & 0.533 \\
        CountryOfficialLanguage   & raw BERT-large-cased & 0.936 & 0.749 & 0.797 \\
        PersonCauseOfDeath        & raw BERT-large-cased & 1.000 & 0.500 & 0.500 \\
        PersonEmployer            & raw BERT-large-cased & 0.022 & 0.035 & 0.025 \\
        PersonInstrument          & MLM $c=0$ & 0.820 & 0.603 & 0.580 \\
        PersonLanguage            & MLM $c=1$ & 0.810 & 0.752 & 0.742 \\
        PersonPlaceOfDeath        & raw BERT-large-cased & 1.000 & 0.500 & 0.500 \\
        PersonProfession          & MLM $c=0$ & 0.620 & 0.540 & 0.538 \\
        RiverBasinsCountry        & MLM silver separate & 0.647 & 0.628 & 0.584 \\
        StateSharesBorderState    & MLM silver separate & 0.280 & 0.184 & 0.198 \\\hline
        \textbf{Average}             & NA & 0.745 & 0.547 & 0.543 \\\hline
    \end{tabular}
    \label{tab:mlm_joint}
\end{table}

As illustrated in Table \ref{tab:mlm_1}, \ref{tab:mlm_2}, we find mixed results from our initial experimentation with task-specific pre-training: the performance dramatically improves with some relations, and dramatically drops on others. The trend is consistent across different configurations of intermediate pre-training, while the exact values slightly differ across different window sizes $c$ with no dominant configuration. This implies, for knowledge graph population with BERT language models, one size doesn't fit all; we need separately-fine-tuned LM checkpoints for different relations to for the best results. Adapting BERT to attend to different relations separately would be impractical with only the challenge training set because of the small sizes of training data; however, with the much larger silver datasets retrieved from Wikidata, we are now able to elicit a family of BERT checkpoints, each dedicated to one or a few relations, where different checkpoints are reminded of different types of factual knowledge. When jointly used for link prediction, our family of BERT checkpoints exhibit superior performance over any single BERT model, as shown in Table \ref{tab:mlm_joint}.

As an additional comment, we present the sizes of our additional MLM training data in Table \ref{tab:mlm_sizes}, where we show that there is not a clear dependency between the sizes of the training sets and the performance of task-specifically pre-trained checkpoints. This means, the discrepancy in performance of this MLM training is not strongly related to the sizes, but rather related to the properties of the knowledge required for each relation.

\section{Prompt Decomposition for Improved Candidate Generation}
\label{Sec:candidate_generation}

In this section, we discuss the approaches explored for generating better candidate objects by prompt-based link-prediction. Our efforts here can be broadly classified into two categories: using better prompts and decomposing the prompts.

\subsection{Prompt Elicitation}
\label{sec:CG_prompt_elicitation}

On the elicitation of better prompts, we experimented with both manual and automatic approaches. For the relation \textit{PersonInstrument}, in order to help BERT ground the names to the corresponding musicians, we explicitly provide the entity type ``musician'' as part of the prompt: ``\textcolor{red}{The musician} [SUBJ] plays [OBJ], which is an instrument''. For the relation \textit{PersonEmployer}, we simplify the prompt into a concise sentence to the same effect: ``[SUBJ] works at [OBJ]''. 

For automatic elicitation of better prompts, we follow \cite{jiang_how_2020} in retrieving sentences from Wikipedia as potential prompts. 
First, we split the Wikipedia passages into sentences. Then we check each sentence with all the subject-object pairs in the LM-KBC dataset to identify if the sentence contains both entities with exact text matches under lowercase. For generating prompts from the selected sentences, we follow the mining-based generation methods introduced by \cite{jiang_how_2020}, which includes elicitation of middle-word prompts and dependency-based prompts. The middle-word prompts are generated by retaining words in between the subject-object pair. The dependency-based prompts are generated based on the dependency tree, where the prompt border are selected based on the left-most and right-most word of the shortest dependency path between the two entities. 


In evaluation, we take the top-20 most frequent prompts in the challenge training set, and rank the performance of BERT LM on the training set using each of these retrieved prompts. Finally, the average of the top-performing prompts is used as the prediction score for each predicted object entity. We iteratively add more prompts to the average in the order of the ranking, and take the best-performing combination (on the training set of the challenge) that is at least 1\% higher than the previous-best using less prompts.\footnote{The 1\% margin is introduced to prevent overfitting.}

Results for this experiment are displayed in Table \ref{tab:mlm_multiprompt}. Contrary to results reported in \cite{jiang_how_2020}, by using a very similar approach over a similar set of relations, we are not observing the same scale of improvement with increasing number of prompts involved; in fact, most of the times the best F-1 score is reached with 1 prompt template, which often is the manually-written one. We argue that this difference is due to the difference in evaluation metrics: we care about the F-1 scores rather than the macro average accuracies, which attaches higher importance to the precision of methods.

\begin{table}[h!]
    \centering
    \caption{\textcolor{blue}{Dev set results for all relations;} on top of the results from Table \ref{tab:mlm_joint}, uses average of top-performing prompts.}
    \begin{tabular}{|c|c|c|c|c|}
        \hline
        \multirow{2}{*}{Relation} & \multirow{2}{*}{Number of Prompts} & \multicolumn{3}{|c|}{Result} \\\cline{3-5}
         & & Precision & Recall & F-1 \\\hline
        ChemicalCompoundElement   & 1 & 0.897 & 0.833 & 0.839 \\
        CompanyParentOrganization & 1 & 1.000 & 0.680 & 0.680 \\
        CountryBordersWithCountry & 2 & 0.653 & 0.580 & 0.548 \\
        CountryOfficialLanguage   & 1 & 0.894 & 0.789 & 0.812 \\
        PersonCauseOfDeath        & 1 & 1.000 & 0.500 & 0.500 \\
        PersonEmployer            & 1 & 0.028 & 0.065 & 0.036 \\
        PersonInstrument          & 1 & 0.820 & 0.634 & 0.617 \\
        PersonLanguage            & 1 & 0.820 & 0.752 & 0.745 \\
        PersonPlaceOfDeath        & 1 & 1.000 & 0.500 & 0.500 \\
        PersonProfession          & 1 & 0.586 & 0.600 & 0.549 \\
        RiverBasinsCountry        & 1 & 0.647 & 0.628 & 0.584 \\
        StateSharesBorderState    & 1 & 0.280 & 0.184 & 0.198 \\\hline
        \textbf{Average}          & NA & 0.719 & 0.562 & 0.545 \\\hline
    \end{tabular}
    \label{tab:mlm_multiprompt}
\end{table}

\subsection{Prompt Decomposition}
\label{sec:CG_prompt_decomposition}

One issue we found hindering the performance of the baseline is, for the relation \textit{StateSharesBorderState}, the subjects and objects are in fact not always states. For instance, the subject ``Andalusia'' is a autonomous community, and the subject ``Hebei'' is a province. By calling these subjects ``states'', as the baseline prompt does, BERT gets confused and outputs irrelevant object entities. To address this issue, we introduce a pre-condition prompt asking which kind of location the subject is, with the prompt template ``[SUBJ], as a place, is a [MASK].'' From all the candidate tokens that BERT generates for the mask, we look for the following set of keywords: [state, province, department, city, region]. The top-ranked keyword will be taken as the type of the subject. Then, the formal prompt would go like ``[SUBJ] [KEYWORD] shares border with [MASK] [KEYWORD]''. In our experiments, we observe a positive effect from this amendment: the F-1 score for the relation ``StateSharesBorderState'' is increased from 0.112 in baseline to 0.162 just with this one change in prompt formulation.

\section{Explorations for Candidate Selection}
\label{Sec:candidate_selection}

In this section, we discuss the approaches explored for selecting appropriate candidates from the distribution of candidate tokens as outputted above. Following the baseline, we consider only the top-100 candidates for each object-to-predict.
One immediate observation from the baseline is, among the top predicted tokens, there often exists pronouns, such as \textit{me, them, it}, or determiners, such as \textit{the, a, some}. Thus, we remove these pronouns and determiners as a post-processing step to clean up the results.

Another observation from the baseline is, that the default threshold of 0.5 is too harsh for many relations, and could be relaxed to optimize the F-1 scores. To this end, we exhaustively search the thresholds between 0 and 0.95 by steps of 0.01, and select the best thresholds based on training set F-1.

\begin{table}[h!]
    \centering
    \caption{Development set results for all relations; comparison is between the original baseline (where $t=0.5$) and the baseline after applying adaptive thresholds (where $t=xx$ are listed in the last column).}
    \begin{tabular}{|c|c|c|c|c|c|c|c|}
        \hline
        \multirow{2}{*}{Relation} & \multicolumn{3}{|c|}{Baseline $t=0.5$} & \multicolumn{4}{|c|}{Baseline $t=xx$} \\\cline{2-8}
         & Precision & Recall & F-1 & Precision & Recall & F-1 & $t=$ \\\hline
        ChemicalCompoundElement   & 0.960 & 0.060 & 0.083 & 0.612 & 0.534 & 0.538 & 0.06 \\
        CompanyParentOrganization & 0.960 & 0.680 & 0.680 & 1.000 & 0.680 & 0.680 & 0.99 \\
        CountryBordersWithCountry & 1.000 & 0.087 & 0.122 & 0.624 & 0.528 & 0.533 & 0.05 \\
        CountryOfficialLanguage   & 0.957 & 0.703 & 0.752 & 0.936 & 0.749 & 0.797 & 0.32 \\
        PersonCauseOfDeath        & 0.880 & 0.520 & 0.420 & 1.000 & 0.500 & 0.500 & 0.99 \\
        PersonEmployer            & 1.000 & 0.000 & 0.000 & 0.022 & 0.035 & 0.025 & 0.03 \\
        PersonInstrument          & 1.000 & 0.340 & 0.340 & 0.560 & 0.504 & 0.509 & 0.42 \\
        PersonLanguage            & 0.900 & 0.412 & 0.431 & 0.740 & 0.671 & 0.670 & 0.15 \\
        PersonPlaceOfDeath        & 0.980 & 0.500 & 0.500 & 1.000 & 0.500 & 0.500 & 0.99 \\
        PersonProfession          & 1.000 & 0.000 & 0.000 & 0.100 & 0.263 & 0.131 & 0.04 \\
        RiverBasinsCountry        & 0.960 & 0.342 & 0.381 & 0.626 & 0.560 & 0.547 & 0.05 \\
        StateSharesBorderState    & 0.900 & 0.000 & 0.000 & 0.110 & 0.130 & 0.112 & 0.05 \\\hline
        \textbf{Average}          & 0.958 & 0.304 & 0.309 & 0.488 & 0.455 & 0.447 & NA \\\hline
    \end{tabular}
    \label{tab:adaptive_thres}
\end{table}

Apart from the above, there can be different numbers of answer objects for different subject-relation pairs, and we notice that for those entries with larger numbers of answers, generally there are more candidate objects with a substantial normalized prediction score. This is in contradiction to the assumption behind normalized prediction scores: normalized prediction scores are contemplated as distributions over the tokens, where there is supposed to be only one true answer. But when there are multiple confidently-predicted candidates, the prediction score of each of them is diluted. Therefore, by setting a common threshold to all entries, the answers for those entries with more true answers are disadvantaged, and have a larger chance of being missed out.

To mitigate this effect, we first tried removing Softmax from the BERT MLM prediction head. The idea is, by removing the normalization and exposing the raw scores, all confident predictions should receive high scores and thus can be thresholded equally. However, experiment results show that by removing the Softmax function performance drops consistently across all relations. We speculate that this is because the range of raw prediction scores vary from sentence to sentence, without a normalization operation the scores themselves are too noisy.

\begin{table}[h!]
    \centering
    \caption{Development set results for all relations with sticky thresholds; results are to be compared with those in Table \ref{tab:mlm_joint}.}
    \begin{tabular}{|c|c|c|c|c|}
        \hline
        \multirow{2}{*}{Relation} & \multirow{2}{*}{Sticky Ratios} & \multicolumn{3}{|c|}{Sticky Ratios} \\\cline{3-5}
         & & Precision & Recall & F-1 \\\hline
        ChemicalCompoundElement   & null & 0.897 & 0.833 & 0.839 \\
        CompanyParentOrganization & null & 1.000 & 0.680 & 0.680  \\
        CountryBordersWithCountry & 0.4 & 0.688 & 0.549 & 0.533  \\
        CountryOfficialLanguage   & 0.91 & 0.936 & 0.749 & 0.797  \\
        PersonCauseOfDeath        & null & 1.000 & 0.500 & 0.500 \\
        PersonEmployer            & 0.76 & 0.522 & 0.035 & 0.025  \\
        PersonInstrument          & 0.43 & 0.813 & 0.648 & 0.607 \\
        PersonLanguage            & null & 0.810 & 0.752 & 0.742 \\
        PersonPlaceOfDeath        & null & 1.000 & 0.500 & 0.500 \\
        PersonProfession          & 0.49 & 0.617 & 0.543 & 0.539 \\
        RiverBasinsCountry        & 0.85 & 0.647 & 0.628 & 0.584 \\
        StateSharesBorderState    & 0.64 & 0.274 & 0.190 & 0.202 \\\hline
        \textbf{Average}          & NA & 0.767 & 0.551 & 0.546 \\\hline
    \end{tabular}
    \label{tab:sticky}
\end{table}

We further tried keeping the Softmax layer, but additionally introducing sticky thresholds. That is, we rank the candidate objects by prediction scores and iterate over them, when a candidate object does not have enough prediction score to meet the threshold, but is relatively close to its previous candidate (for instance, $> 80\%$ the prediction score of its previous candidiate), we accept this candidate as well. We search for optimal sticky ratios along with the thresholds. Unfortunately, we observe that while for many relations the best F-1 score is reached with non-empty sticky ratios, only a very slight improvement is achieved, as shown in Table \ref{tab:sticky}.


\section{Final Results}
\label{Sec:experiments}

\begin{table}[h]
    \centering
    \caption{A Summary of \textbf{Dev} set results of the baseline and incrementally applying various techniques.}
    \begin{tabular}{|c|c|c|c|}
    \hline
    Method & Precision & Recall & F-1 \\\hline
    Baseline & 0.958 & 0.304 & 0.309 \\
    + threshold tuning & 0.488 & 0.455 & 0.447 \\
    + prompt decomposition & 0.510 & 0.469 & 0.464 \\
    + MLM training & 0.745 & 0.547 & 0.543 \\
    + sticky ratios & 0.767 & 0.551 & 0.546 \\\hline
    \end{tabular}
    \label{tab:all_devs_ablation}
\end{table}

\begin{table}[h!]
    \centering
    \caption{Test set results for all relations; between the baseline method and our final results.}
    \begin{tabular}{|c|c|c|c|c|c|c|}
        \hline
        \multirow{2}{*}{Relation} & \multicolumn{3}{|c|}{Baseline} & \multicolumn{3}{|c|}{Ours} \\\cline{2-7}
         & Precision & Recall & F-1 & Precision & Recall & F-1 \\\hline
        ChemicalCompoundElement   & 0.980 & 0.069 & 0.098 & 0.835 & 0.843 & 0.817 \\
        CompanyParentOrganization & 0.900 & 0.740 & 0.640 & 1.000 & 0.740 & 0.740 \\
        CountryBordersWithCountry & 0.980 & 0.105 & 0.119 & 0.571 & 0.667 & 0.557 \\
        CountryOfficialLanguage   & 0.980 & 0.719 & 0.786 & 0.960 & 0.749 & 0.803 \\
        PersonCauseOfDeath        & 0.860 & 0.500 & 0.360 & 1.000 & 0.500 & 0.500 \\
        PersonEmployer            & 0.480 & 0.020 & 0.020 & 0.480 & 0.020 & 0.020 \\
        PersonInstrument          & 1.000 & 0.360 & 0.360 & 0.840 & 0.551 & 0.483 \\
        PersonLanguage            & 1.000 & 0.376 & 0.428 & 0.897 & 0.748 & 0.773 \\
        PersonPlaceOfDeath        & 0.980 & 0.500 & 0.480 & 1.000 & 0.500 & 0.500 \\
        PersonProfession          & 1.000 & 0.000 & 0.000 & 0.689 & 0.658 & 0.620 \\
        RiverBasinsCountry        & 0.960 & 0.404 & 0.429 & 0.529 & 0.562 & 0.496 \\
        StateSharesBorderState    & 0.900 & 0.006 & 0.010 & 0.385 & 0.252 & 0.291 \\\hline
        \textbf{Average}             & 0.960 & 0.319 & 0.311 & 0.765 & 0.566 & 0.550 \\\hline
    \end{tabular}
    \label{tab:final_experiment}
\end{table}

In Table \ref{tab:all_devs_ablation} is a summary of the techniques we've tried. It is not surprising that applying an adaptive threshold scheme brings substantial improvements; on the other hand, it is interesting how task-specific MLM training brings another pronounced boost in performance. Prompt decomposition shows an moderate but convincing improvement, whereas the improvement from adding in the sticky ratios is negligible. 

In Table \ref{tab:final_experiment} are the final results for our system on the challenge test set, as recorded on the leaderboard\footnote{\url{https://codalab.lisn.upsaclay.fr/competitions/5815}}. This final set of results is acquired under the following setup: we use the family of BERT LM checkpoints based on BERT-large-cased, as presented in Table \ref{tab:mlm_joint} in Section \ref{Sec:pretraining}; we use our manually updated set of prompts (for computation speed) as in Section \ref{sec:CG_prompt_elicitation}, with thresholds as assigned in Section \ref{Sec:candidate_selection}, ignoring sticky ratios; we use the type-assignment decomposition for ``StateSharesBorderState''.

\section{Conclusions}

We have explored methods to improve Knowledge Graph population with LMs under the \textbf{track 1} constraint of using BERT as the language model backbone. In particular, we explored improving the LM representation, candidate object generation and candidate selection. We have made significant progress against the baseline method, and have also found remaining issues, which, if addressed, would bring further gain in performance and/or versatility. We highlight the following as promising areas of future work: 1) efficient intermediate fine-tuning for arbitrary relations; 2) automatic prompt decomposition, with more powerful LM backbones; 3) alternative re-ranking methods for independent judgement of candidate validity.

\begin{acknowledgments}
  The authors thank the challenge organizers for their timely and helpful response to inquiries, and the reviewers for their valuable comments. This work is supported in part by a Mozzila Informatic PhD scholarship. 
\end{acknowledgments}

\bibliography{references}

\appendix

\section{Hyper-parameters}

In this section, we discuss the selection of hyper-parameters. For MLM training (\S \ref{Sec:pretraining}), the learning rate is 5e-6, selected from (1e-5, 5e-6, 1e-6); the batch size is 64, which is the largest power of 2 possible for our GPUs. We train for 10 epochs for the challenge training set, and 5 when concatenated with silver data in order to control the overall training steps at a manageable level. Checkpoints are saved at the end of every epoch, where the checkpoint with lowest dev2 set perplexity is considered the best.

\end{document}